\documentclass{article}
\usepackage{arxiv}              

\usepackage[utf8]{inputenc}
\usepackage[T1]{fontenc}
\usepackage{microtype}

\usepackage{graphicx}
\usepackage{amsmath}
\usepackage{amssymb}
\usepackage{booktabs}
\usepackage{float}
\usepackage{xcolor}
\usepackage[numbers]{natbib}
\usepackage[font=small,labelfont=bf,skip=6pt]{caption}

\makeatletter
\renewenvironment{table}
  {\setlength{\abovecaptionskip}{7\p@}%
   \setlength{\belowcaptionskip}{\z@}%
   \@float{table}}
  {\end@float}
\makeatother

\floatstyle{ruled}
\newfloat{algorithm}{tbp}{loa}
\floatname{algorithm}{Algorithm}
\floatstyle{plain}

\definecolor{linknavy}{rgb}{0.10,0.25,0.55}
\definecolor{citegreen}{rgb}{0.10,0.40,0.25}
\usepackage[colorlinks=true, linkcolor=linknavy, citecolor=citegreen,
            urlcolor=linknavy, breaklinks=true]{hyperref}
\usepackage[capitalise]{cleveref}

\newcommand{\etal}{\emph{et al.}}

\renewcommand{\headeright}{A Preprint}
\renewcommand{\shorttitle}{}   
\renewcommand{\undertitle}{Technical Report}

\title{Hard-Region Supervision:\\
\#1 on the Waymo Open Dataset\\
2D Video Panoptic Segmentation Leaderboard}
\author{%
  Jinghan Yang \\
  \texttt{jinghanyang2023@gmail.com} \\
}
\date{}

\begin{document}

\maketitle

\begin{abstract}

We describe our winning entry to the Waymo Open Dataset 2D Video Panoptic Segmentation Challenge. The task asks for a semantic class at every pixel of every frame and, for countable objects, an identity that holds across 100 frames and across five overlapping cameras. We build on DVIS++, a cascade of a segmenter, a tracker, and a refiner, as our baseline. We propose hard region supervision (HRS) to improve the baseline. In particular, we use the baseline to define the hard region as where it makes mistakes, and design a loss and an auxiliary prediction head for this region. The auxiliary head is used only in training and removed at test time, so at inference the model trained with HRS has the same architecture as the baseline. In addition, we propose three test-time steps that further improve the results: a two-model ensemble, a merge of the segmenter's output into the final panoptic map, and cross-camera identity linking. On the challenge test set, our entry reaches 0.3547 wSTQ, 0.2071 wAQ, and 0.6075 mIoU, ranking first on all three metrics. It is 3.6 wSTQ points ahead of the second entry and 2.4 points ahead of our DVIS++ baseline.
\end{abstract}

\section{Introduction and Related Work}
\label{sec:intro}

This report describes our entry to the Waymo Open Dataset 2D Video Panoptic Segmentation Challenge. The task is to assign every pixel of every frame a semantic class and, for countable objects, an instance identity that stays the same across frames and across the five cameras that make up the panoramic view. It is demanding in three separate ways at once: the semantic classes are heavily imbalanced, with ten of the twenty-eight classes together holding only 0.175\% of labeled pixels; test videos are 100 frames long, so identities must survive all 100 frames; and the five cameras overlap, so an object straddling a camera boundary is segmented twice and must be recognized as one object.

We build on DVIS++~\cite{dvisplus}, which decomposes the video panoptic segmentation (VPS) task into a per-frame segmenter, a
tracker and a temporal refiner trained one after another. We change it in two
places. 
First, we find that on WOD-PVPS, a large-scale dataset, the baseline performs poorly on long-tail classes and object boundaries in segmentation. We therefore use the baseline to mark the hard region, where it makes wrong predictions, and propose a boosting-style algorithm with hard region supervision that improves performance on these baseline errors without reducing performance on the pixels the baseline already predicts correctly.
Second, we find that the baseline's decoupled design, with three modules, currently uses a suboptimal inference that hurts both AQ and mIoU. Our proposed inference pipeline improves both metrics.

On the challenge test set, the resulting system reaches a weighted STQ of $0.3547$, a weighted AQ of $0.2071$ and an mIoU of $0.6075$, placing first on the leaderboard. The second-place entry reaches $0.3187$, $0.1781$ and $0.5702$, and the baseline we build upon reaches $0.3308$, $0.1929$ and $0.5675$. The remainder of this section reviews the work this builds on; Section~\ref{sec:method} describes the methods and Section~\ref{sec:experiment} the experiments.

\subsection{Challenges in Semantic Segmentation}
Two failure cases dominate semantic segmentation quality. The first is the object boundary. A mask is mostly interior,
so a prediction that is correct everywhere except at its edge still scores well
under mask IoU; Cheng~\etal~\cite{cheng2021boundaryiou} show this directly,
proposing Boundary IoU because the standard measure is insensitive to exactly
the errors that concentrate at object edges. The second is the long tail. Pixel
counts per class span orders of magnitude---LVIS~\cite{gupta2019lvis} made this
the defining property of a segmentation benchmark---so a loss averaged over
pixels is dominated by the head classes and the tail contributes almost nothing
to the gradient. The two are hard in different ways: boundaries are a small
fraction of every object, while rare classes are a small fraction of the
dataset. Prior work fixes each by changing the loss, the sampling, or the
architecture.

On the boundary side, Kervadec~\etal~\cite{kervadec2019boundary} 
propose a weighted scheme for pixels, where every pixel is weighted by its distance to the boundary rather than uniform weights.
PointRend~\cite{kirillov2020pointrend} concentrates prediction and supervision on the points where the mask is least certain, which in practice could be boundary pixels. Mask2Former~\cite{mask2former} adopts PointRend for sampling. 
Boundary-preserving Mask R-CNN~\cite{cheng2020boundary} adds supervision on the boundary band through a parallel boundary head, whose features are fed into the mask head, so that the mask prediction is built on features trained to be boundary-aware.

On the long-tail side, loss-based methods reshape the gradient across classes: focal loss~\cite{lin2017focal} down-weights well-classified pixels by a factor of $(1-p_t)^{\gamma}$; class-balanced loss~\cite{cui2019classbalanced} reweights each class by the inverse of its effective number of samples; equalization loss~\cite{tan2020eql} removes the negative gradients that
frequent classes impose on rare ones. EQLv2~\cite{tan2021eqlv2}, Seesaw
loss~\cite{wang2021seesaw} and logit adjustment~\cite{menon2021logit} are
variations on the same idea, differing only in the statistic that sets the
correction: the accumulated positive-to-negative gradient ratio, the
cumulative sample ratio, and the log class prior respectively.
Training-paradigm methods change what the model sees instead, by resampling,
by staging the training, or by augmentation:
repeat-factor sampling~\cite{gupta2019lvis} repeats an image in proportion to the rarity of its rarest category; Kang~\etal~\cite{kang2020decoupling} train the representation with instance-balanced sampling and then retrain or rescale only the classifier; copy-paste augmentation~\cite{ghiasi2021copypaste} pastes object instances between images.

\subsection{Video panoptic segmentation.}
Video panoptic segmentation (VPS) asks for a semantic class label at every pixel of
every frame and, for pixels of countable \emph{thing} classes, an instance identity
that stays consistent across frames~\cite{kim2020vps}. Uncountable \emph{stuff}
classes such as road and sky receive a class label only. 
The task therefore subsumes
three problems at once: which class (semantic segmentation), which object (instance
segmentation), and which object over time (tracking). It descends from image panoptic
segmentation, which unified the first two into a single per-pixel prediction of class
plus instance id.

VPS is commonly measured by two metrics, VPQ and STQ. VPQ~\cite{kim2020vps} is panoptic quality computed over tubes of $k$ consecutive frames and averaged over $k$. STQ~\cite{weber2021step} was proposed because VPQ under-rewards long-term tracking, since it only scores short tubes; it separates the two failure modes into a segmentation quality (SQ), measured by mIoU, and an association quality (AQ), and scores a method by their geometric mean. The Waymo Open Dataset~\cite{mei2022wodpvps} adopts a weighted variant, wSTQ, which weights each pixel by the inverse of the number of cameras observing it so that overlapping views are not double-counted; on single-camera data it reduces to STQ. Following the Waymo competition convention, we report wSTQ together with its two components, wAQ and mIoU.

Video segmentation benchmarks split by what they annotate. For video
instance segmentation, YouTube-VIS~\cite{yang2019ytvis} and the heavily
occluded OVIS~\cite{qi2022ovis} label only \emph{thing} tracks;
VSPW~\cite{miao2021vspw} labels every pixel but no identities; and
VIPSeg~\cite{miao2022vipseg} labels both over 3,536 in-the-wild videos,
which makes it the standard video panoptic benchmark---DVIS++, our baseline,
is evaluated on all four. Driving video is annotated separately, and more
sparsely: Cityscapes-VPS and VIPER~\cite{kim2020vps} label every fifth frame,
while KITTI-STEP and MOTChallenge-STEP~\cite{weber2021step} label every frame
of long sequences. The Waymo Open Dataset~\cite{mei2022wodpvps} is the
largest of them by annotated real frames and the only one that is panoramic,
with five cameras rather than one. This report uses its 2D Video Panoptic
Segmentation Challenge split: training sequences carry 5 labelled frames and
evaluation sequences about 100, sampled at 5\,Hz across 5 cameras (3 front at
$1920\times1280$, 2 side at $1920\times886$) over 28 semantic classes.

\paragraph{From dense prediction to set prediction.}
Detection and segmentation problems (DSP) were for a decade solved by dense
prediction: a network scores a grid of hand-designed candidates --- anchors,
proposals or pixels --- and a post-processing step such as non-maximum suppression
selects among them~\cite{ren2015fasterrcnn,he2017maskrcnn}. DETR~\cite{detr} replaced
this with set prediction: a fixed set of learned queries is matched one-to-one with
the ground-truth objects, so each query directly outputs one object, and no candidate
grid or suppression is needed. The consequence is that the query becomes the
universal interface for DSP. Once an object is one query, detection, instance,
semantic, panoptic and video segmentation all collapse into the same formulation ---
MaskFormer~\cite{cheng2021maskformer}, Mask2Former~\cite{mask2former} and their video
Mask2Former~\cite{mask2former} carried the formulation to image
segmentation, unifying semantic, instance and panoptic segmentation in one
architecture --- a backbone extracts image features, a pixel decoder produces a
high-resolution per-pixel embedding together with a multi-scale feature pyramid,
and a transformer decoder refines a set of object queries by attending over that
pyramid. Each query then predicts a class and a mask, the mask being the inner
product between the query's mask embedding and the per-pixel embedding.
MinVIS~\cite{minvis} extends this to video without changing the architecture or the
objective: frames are trained independently as images, and instances are associated
only at inference, by matching query embeddings between consecutive frames. No
tracking is ever trained.

\paragraph{Unified and decoupled video models.}
Early video panoptic methods needed a separate mechanism for things and for stuff;
the query-based reformulation removed that need, since a single set of queries
represents both and can be carried across frames, as in Video
K-Net~\cite{li2022videoknet} and TubeFormer-DeepLab~\cite{kim2022tubeformer}.
Recent work splits into two lines: unified models that treat video panoptic,
instance and semantic segmentation as one query-propagation
problem~\cite{athar2023tarvis,li2023tubelink,li2024omgseg}, and decoupled models
that freeze a per-frame segmenter and train a tracker on its
queries~\cite{minvis,dvis,dvisplus}. The latter currently define the state of the
art, and are the line we build on.

\section{Method}
\label{sec:method}
We take DVIS++~\cite{dvisplus} as our baseline, described in section~\ref{subsec:baseline}, and build on it in two places.
We propose hard-region supervision (HRS), a boosting-style scheme: a trained model marks the pixels it still gets wrong, and a second model is trained with an additional loss on those pixels. Section~\ref{subsec:hard-region} defines the hard region (Equation~\ref{eq:hard-region}), the hard region loss (Equation~\ref{eq:loss-propose}), and the $\beta$-head through which this loss is mainly applied.
At inference, 
we propose decoupled inference merge (DIM), a test-time fusion: the cascade
produces its panoptic map from the tracker and refiner alone, discarding the
segmenter's own prediction, and DIM puts it back.
Equation~\ref{eq:dim} defines the merge and
section~\ref{sec:inference} the pipeline it sits in; the same stage
reconciles instance identities across the five cameras
(Algorithm~\ref{alg:cross-cam}).

\subsection{Baseline}
\label{subsec:baseline}
We take DVIS++~\cite{dvisplus} as our baseline. It is a cascade of three
modules: a segmenter $f$ (a backbone and a Mask2Former~\cite{mask2former} head) that predicts
per-frame masks and classes, a referring tracker $g$ that associates them
across frames, and a temporal refiner $h$ that produces the final masks for the video. 

\paragraph{Loss and cascaded training.}
\label{par:cascade-flaw}
The baseline uses the Mask2Former objective~\cite{mask2former} unchanged. A
Hungarian matcher~\cite{detr} assigns each predicted query to at most one ground-truth
object, and the matched pairs are supervised by
\begin{equation}
\mathcal{L}_{\text{baseline}}
= \lambda_1 \mathcal{L}_{\text{Dice}}
+ \lambda_2 \mathcal{L}_{\text{BCE}}
+ \lambda_3 \mathcal{L}_{\text{CE}},
\label{eq:baseline}
\end{equation}
where $\mathcal{L}_{\text{Dice}}$ and $\mathcal{L}_{\text{BCE}}$ supervise
each matched mask and $\mathcal{L}_{\text{CE}}$ is the query classification
loss with a down-weighted no-object class. The same loss is applied at every
decoder layer as an auxiliary loss. 

The two mask terms are not evaluated densely. Following
PointRend~\cite{kirillov2020pointrend}, each matched mask is scored on $P$
sampled points: $R \cdot P$ candidates are drawn uniformly over the whole
image, with $R > 1$ an oversampling multiplier; they are ranked by an
uncertainty computed from the predicted logits, the most uncertain $0.75P$
are kept, and the remaining $0.25P$ are drawn uniformly at random.

The three modules are trained in sequence, each with the earlier stages
frozen. The segmenter $f$ is trained first, as a per-frame instance
segmentation model. The tracker $g$ is trained on top of the frozen $f$ and
gives each object an identity that persists across frames, fusing the
current frame's queries with the previous frame's---the online setting. The
refiner $h$ is trained last, on top of both frozen modules, and lets each
object exchange information with the full temporal horizon---the offline
setting.
We refer to this sequential scheme as \emph{cascaded training} in this technical report. 
What the cascade passes along is queries. The queries of $g$ are formed only from the queries of $f$, and those of $h$ only from those of $g$ and $f$; the mask features produced by $f$ are shared by all three, but only as the frozen map from which each module's mask head decodes its masks ($g$ adds a learned $1\times1$ projection). Pixel features are never consulted while $g$ and $h$ form their queries---they enter once, at the very end, to turn a finished query into a mask. 

\paragraph{Inference modes.}
DVIS++ turns the same set of object queries into a prediction in two different ways, and the choice decides what the output can contain: one obtains a dense semantic map, and the other obtains a panoptic prediction with associated instance IDs.

\emph{Video panoptic segmentation (VPS)} selects. A query survives only if
its top class is not the no-object slot and its confidence exceeds a
threshold $\tau$; each surviving mask is scaled by that confidence and every
pixel goes to the single highest-scoring query,
\begin{equation}
i^\star(x) \;=\; \arg\max_q \; c_q \, m_q(x),
\qquad
c_q = \max_k p_q(k),
\end{equation}
provided that query's own mask also exceeds $\tau$ there. Small segments are
dropped and stuff segments sharing a class are merged. Every pixel gets a
class and an instance identity---or nothing, if no retained query claims it.

\emph{Video semantic segmentation (VSS)} sums instead of selecting. With the
no-object column dropped, every query contributes to every pixel,
\begin{equation}
p(c \mid x) \;=\; \sum_q p_q(c)\, m_q(x),
\qquad
\text{label}(x) \;=\; \arg\max_c \; p(c \mid x).
\end{equation}
No threshold is applied, so every pixel receives a class, with no instance
identity and no possible abstention.

\subsection{Proposed}
We make two changes to the baseline: one to how it is trained, and one to how it is run.

The first is that training treats every pixel alike, except for the uncertainty-based point sampling of PointRend~\cite{kirillov2020pointrend}, which samples uncertain points more densely for the loss. However, we empirically find that the trained model still performs poorly on boundaries and rare classes.
We take a boosting view for solving this. 
A first model is trained to convergence following the baseline setup exactly; the pixels it still gets wrong are identified
and called the \emph{hard region}; and a second model is trained with an
additional loss on exactly those pixels, carried by a prediction branch that
is discarded afterwards. The two are combined at test time, so the second
model is asked not to be better on its own but to be right where the first is
wrong. Section~\ref{subsec:hard-region} describes this.

The second change follows from the cascaded objectives and training design. In practice, we find that on WOD-PVPS, the refiner's panoptic output drops thing-pixels that the segmenter had already detected correctly. We attribute this to the cascade: each module in the baseline is trained on the frozen output of the one before it and forms its queries only from upstream queries, never from the pixel features, so information the segmenter held can be lost by the time the refiner produces the final map. Therefore rather than discarding the segmentation from $f$ and taking the prediction from the tracking modules alone, we recover these pixels at inference by also merging in the segmenter's output. Section~\ref{ssub:inference} describes that merge, together with the cross-camera alignment algorithm that associates instance IDs across the five cameras.

\subsubsection{Hard-Region Supervision (HRS)}
\label{subsec:hard-region}
Hard region supervision (HRS) is a boosting-style scheme~\cite{freund1997adaboost}: it trains a second model, the student $\phi_s$, under the guidance of a trained model, the teacher $\phi_t$, which marks where the hard pixels are, and trains the student to reduce its errors on these regions. At inference, the student $\phi_s$ can be used alone for panoptic segmentation, or the two models can be ensembled for semantic segmentation (Section~\ref{ssub:inference}).

\paragraph{Loss and Model.}
Given a teacher model $\phi_t$, a query-based instance segmenter that
predicts a set of queries, a matching algorithm assigns the predicted
queries to the ground-truth objects one to one. For a matched pair, with
ground-truth mask $y$ and the teacher's predicted mask $m_t$ for the same
object, the raw hard region is
\begin{equation}
H = \{\, x : m_t(x) \neq y(x) \,\}.
\label{eq:hard-region}
\end{equation}

$H$ contains only the misclassified pixels. These are typically sparse and
thin, forming a narrow band of one to a few pixels along the object
boundary, and they lie mostly inside the object, so $H$ alone contains few
background pixels. Supervising directly on such an isolated set has two
problems. First, it provides little spatial context, so the features
learned there tend not to generalize. Second, since $H$ is mostly interior,
the hard-region loss admits a degenerate solution: predicting every point
in $H$ as foreground.
We therefore dilate the hard region
by a band of $r$ pixels around its boundary, where $r$ is a hyperparameter
(we use $r=5$); the thickened region $H_r = H \oplus B_r$ keeps its focus on
the hard pixels while supplying enough surrounding context.

We propose adding a hard-region loss for object mask formation, which has the same definition as the Dice and BCE mask losses but is computed only on the hard region. In particular, $\phi_s$ is trained with $\mathcal{L}_{\text{baseline}}$ on its mask head $\alpha$, together with an additional loss defined only on the hard region and mainly supervised through the $\beta$-head (defined below):

\begin{equation}
\mathcal{L}
= \mathcal{L}_{\text{baseline}}(\alpha)
+ \mathcal{L}^{\mathrm{hard}}(\beta)
+ \epsilon\, \mathcal{L}^{\mathrm{hard}}(\alpha),
\qquad
\mathcal{L}^{\mathrm{hard}} = \lambda_4\, \mathcal{L}^{\mathrm{hard}}_{\mathrm{BCE}} + \lambda_5\, \mathcal{L}^{\mathrm{hard}}_{\mathrm{Dice}},
\quad \epsilon \ll 1.
\label{eq:loss-propose}
\end{equation}

\emph{Hard-Region Sampling.}
Like the baseline mask loss, the hard-region
loss is evaluated on sampled points, but the sampling differs from
uncertainty-based sampling in two ways. 
First, the points are
drawn from a restricted region rather than the whole image: the support is
the hard region itself, taken as the disagreement between the teacher's
mask and the ground truth, together with the nearby background pixels lying
within a window of radius $r_{\mathrm{win}}$ of that region but outside the
object. Including this background collar keeps negative examples in the
sample and prevents the $\beta$-head from collapsing to an all-positive
prediction inside the band. Second, the uncertainty step is dropped
entirely: points are drawn uniformly from that support, with no oversampling
and no scoring, because the region has already been selected by the
$\phi_t$'s errors and there is no need to search for uncertain locations
inside it. The budget is smaller than the baseline's, $P/\kappa$ points per mask with
$\kappa > 1$, since the support is a small fraction of
the image.

\emph{$\beta$-head.}
This head can be added to any of the transformer decoders in $f$, $g$, and $h$. Each of these decoders turns a set of queries into per-object predictions through two prediction heads: a class head and a mask head. The mask head, in particular, reads the query after each decoder layer, maps it through an MLP to a mask embedding, and decodes it against the per-pixel embeddings by a dot product to produce the object's mask. We refer to this original mask head as the $\alpha$-head and add a second one, the $\beta$-head, beside it. $\mathcal{L}_{\mathrm{BCE}}$ and $\mathcal{L}_{\mathrm{Dice}}$ are attached to the $\alpha$-head, and $\mathcal{L}^{\mathrm{hard}}_{\mathrm{BCE}}$ and $\mathcal{L}^{\mathrm{hard}}_{\mathrm{Dice}}$ mainly to the $\beta$-head, with only a small weight $\epsilon \ll 1$ on the $\alpha$-head (Equation~\ref{eq:loss-propose}); since the two heads share the decoder, the hard-region loss reaches the queries mainly through $\beta$.


\subsubsection{Inference}
\label{ssub:inference}
Everything in this subsection acts at test time only. We first ensemble the semantic prediction over $K=2$ models, next merge the segmenter's instance map back into the refiner's panoptic map to recover the pixels the cascade drops, and lastly reconcile instance identities across the five cameras.

\paragraph{Segmentation Ensembling (SE)}
The semantic branch is ensembled over $K$ models at test time. Each of the $K$ models is run independently, and their per-pixel class posteriors are averaged uniformly,
\begin{equation}
p_{\mathrm{ens}}(c \mid x) \;=\; \frac{1}{K}\sum_{m=1}^{K} p_m(c \mid x).
\end{equation}
In practice, we use $K=2$: the two models $f_A$ and $f_B$ obtained from the two training rounds.

\paragraph{Decoupled Inference Merge (DIM)}
\label{sec:inference}
We are given the predictions from two modules, $y_f$ and $y_h$, where $y_f$ is a
semantic segmentation map and $y_h$ is a panoptic segmentation map. For stuff
pixels we take the result from $y_f$. For thing pixels, let $L$ be the pixels
detected as thing in $y_h$, which therefore carry an associated instance id, and let
$U$ be the pixels detected as thing in $y_f$ but not in $y_h$.
We label $U$ with a simple KNN, applied independently within each frame. For each
$x \in U$,
\begin{equation}
\text{id}(x) = \text{id}(y^\star),
\qquad
y^\star = \arg\min_{y \in L} \|x - y\|,
\label{eq:dim}
\end{equation}
so $x$ inherits the id of its nearest neighbor in $L$ by pixel distance, while
stuff-pixels keep only their class. Despite its simplicity, this merge improves
both wAQ and mIoU (Table~\ref{tab:official-val}).

\paragraph{Multi-Camera Alignment}
The five cameras overlap at their edges, so a car near the boundary between two
of them is segmented twice, once in each view, and is given an unrelated instance
id in each. Because AQ rewards an object keeping one identity wherever it
appears, these per-camera identities have to be reconciled. Algorithm
\ref{alg:cross-cam} does this on the four adjacent camera pairs. 
Given an adjacent camera pair $(A, B)$, we first estimate a homography $H$
from the two images, which aligns their overlapping area pixel to pixel.
We then warp $A$'s instance prediction into $B$'s view through $H$ and link
an object pair if it satisfies two constraints, same class and sufficient
mask overlap; otherwise the pair is left unlinked. Geometry decides
\emph{where} the two views coincide; the predicted masks decide \emph{what}
is the same object. The two stages use disjoint inputs: $H$ is estimated
from the images alone and never sees a prediction, while the linking uses
only predictions and never sees the pixels. A wrong prediction cannot
corrupt the geometry, and a wrong warp can only cause a missed link, not a
wrong one. In practice, replacing the estimated cross-camera links with the
ground-truth links changes the result only marginally, so
Algorithm~\ref{alg:cross-cam} is close to an oracle alignment\ifdefined\NOAPPENDIX.\else; qualitative
examples are shown in Appendix~\ref{app:cross-cam}.\fi

\begin{algorithm}[H]
\caption{Cross-camera alignment of overlapping objects}
\label{alg:cross-cam}
\noindent\textbf{Input:} images $I_c$ and predicted maps $P_c$ for the five cameras;
the four adjacent pairs $\mathcal{P}$; pixel threshold $\tau$.\\
\noindent\textbf{Output:} one global id per physical object, shared across every camera it spans.
\vspace{4pt}\hrule\vspace{4pt}
\begin{tabbing}
\hspace{2.2em}\=\hspace{1.4em}\=\hspace{1.4em}\=\hspace{1.4em}\=\kill
\textbf{1:} \> initialise a union--find structure $\mathcal{U}$ over all thing instances \\
\textbf{2:} \> \textbf{for each} adjacent pair $(A,B) \in \mathcal{P}$ \textbf{do} \\
\textbf{3:} \> \> $H \gets \textsc{Homography}\big(\textsc{OrbMatch}(I_A, I_B)\big)$
                 \quad \textit{// geometry, from RGB only} \\
\textbf{4:} \> \> $P_A' \gets \textsc{Warp}(P_A, H)$
                 \quad \textit{// $A$'s predicted map, now in $B$'s view} \\
\textbf{5:} \> \> \textbf{for each} thing instance $a$ in $P_A'$ \textbf{do} \\
\textbf{6:} \> \> \> \textbf{for each} thing instance $b$ in $P_B$ \textbf{do} \\
\textbf{7:} \> \> \> \> \textbf{if} $\mathrm{class}(a) = \mathrm{class}(b)$ \textbf{and}
                       $|\mathrm{mask}(a) \cap \mathrm{mask}(b)| \ge \tau$ \textbf{then} \\
\textbf{8:} \> \> \> \> \quad $\mathcal{U}.\textsc{Union}(a, b)$
                       \quad \textit{// same physical object} \\
\textbf{9:} \> \textbf{for each} group $g$ in $\mathcal{U}$ \textbf{do} \\
\textbf{10:} \> \> assign one global id to every instance in $g$ \\
\end{tabbing}
\vspace{-8pt}
\end{algorithm}

\section{Experiment}
\label{sec:experiment}
We evaluate on WOD-PVPS, the five-camera panoramic benchmark the challenge is run on, and report the official wSTQ, wAQ and mIoU. Sections~\ref{subsec:dataset} and~\ref{subsec:models} introduce the dataset details and the experimental setup, including the parameters used to obtain the empirical results; Section~\ref{subsec:results} gives the challenge result and then separates out what each component of Section~\ref{sec:method} contributes to it.

\subsection{Dataset}
\label{subsec:dataset}
We evaluate on the Waymo Open Dataset: Panoramic Video Panoptic Segmentation
(WOD-PVPS). WOD-PVPS~\cite{mei2022wodpvps} provides high-resolution frames across 5 cameras. The 3 front camera have the resolution of 1920x1280, the 2 side cameras have the resolution of 1920x886. 
The split is 70k training / 10k validation / 20k test
images; validation and test are long 100-frame sequences (5\,Hz) for long-term
tracking. 
The training dataset consists of video clips of five frames sampled at 5\,Hz, each from a single camera.
We train on these train video clips, and the model predicts 5 consecutive frames, and use a stitching algorithm for stitching these videos clips to be the 100 frames long videos for the validation and test videos. 

\begin{figure}[htbp]
\centering
\includegraphics[width=\linewidth]{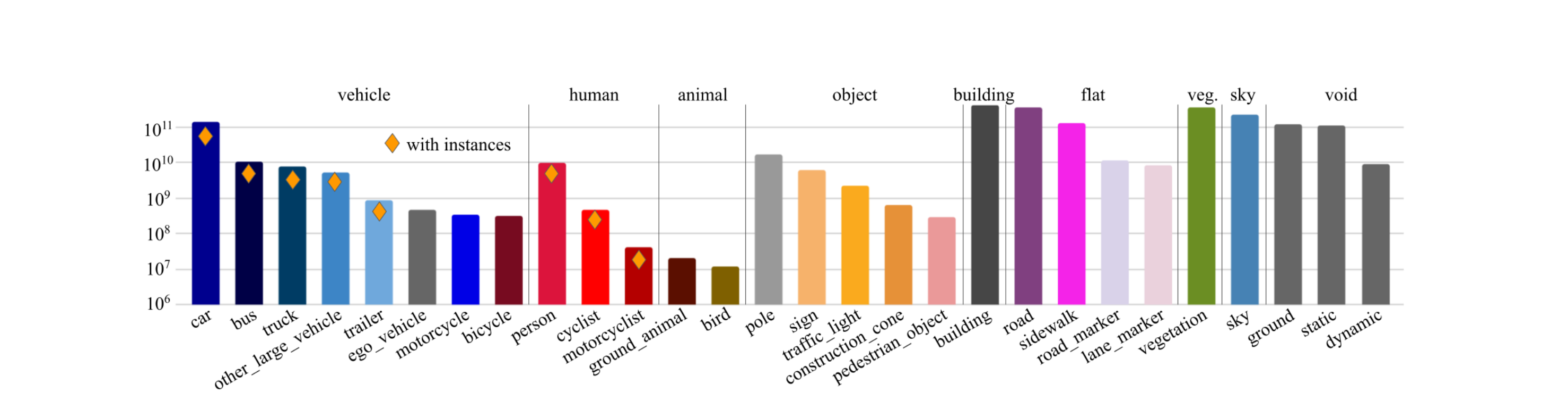}
\caption{Pixel distribution of the 28 semantic categories, reproduced from
Figure~2 of Mei \emph{et al.}~\cite{mei2022wodpvps}. The vertical axis is
the number of pixels per class on a logarithmic scale; classes are grouped
by super-class, and diamonds mark the classes for which instance IDs are
provided.}
\label{fig:mei-class-distribution}
\end{figure}

WOD-PVPS has 28 semantic classes. Figure~\ref{fig:mei-class-distribution}
shows the pixel distribution over these 28 classes and marks which of them
are things and which are stuff. The class distribution is heavily
long-tailed: on the log scale, eighteen classes lie above roughly $10^9$
pixels and ten lie well below it. We denote these ten the \emph{rare}
classes $\mathcal{R}$: trailer, ego vehicle, motorcycle, bicycle, cyclist,
motorcyclist, ground animal, bird, construction cone, and object dragged by
pedestrian. Typically, the rare classes are harder than the common classes,
since they contribute few training pixels. We therefore evaluate on all
classes, and additionally report separate results on $\mathcal{R}$ and on
the remaining common classes.

\subsection{Models and Training}
\label{subsec:models}
We use Mask2Former as the segmenter, with a DINOv2~\cite{oquab2024dinov2} ViT-L/14 backbone wrapped
in a ViT-Adapter~\cite{chen2023vitadapter}. The ViT weights are frozen and only the adapter, pixel
decoder and transformer decoder are trained, which keeps the large backbone
affordable at our batch size. The transformer decoder has nine layers with
hidden dimension $256$ and $200$ object queries over $29$ classes ($28$
classes and one no-object class), and the mask losses are evaluated on
$12{,}544$ PointRend-sampled points per mask with an oversampling ratio of
$3.0$ and an importance-sampling ratio of $0.75$. The matching and loss
weights are $2.0$ for classification and $5.0$ each for mask BCE and Dice,
with a no-object weight of $0.1$. Training uses AdamW at a base learning
rate of $10^{-4}$ with weight decay $10^{-4}$.

The three front cameras are $1280 \times 1920$ while the two side cameras
are $886 \times 1920$, so the resize range is set per camera group: the
shorter side is sampled from $\{960, 1120, 1280\}$ for the front cameras and
from $\{775, 886\}$ for the side cameras, keeping the effective scale
comparable across the five views. Clips are then randomly cropped to
$704 \times 1248$, matching the crop size used at inference, and randomly
flipped.
We follow the cascaded training of DVIS and DVIS++~\cite{dvis, dvisplus} and train the three modules $f$, $g$, and $h$ in sequence, each stage freezing everything before it. These three modules are trained with the same loss definition; however, their objectives differ in how the queries are matched to the ground truth. In $f$ training, the matching is at the frame level: $f$ runs per frame with no temporal modeling, and a clip of shape $(T, B, C, H, W)$ is folded to $(T{\cdot}B, C, H, W)$, so time is only a batch dimension. In $g$ and $h$ training, however, the matching is at the video level. 
Therefore, for a given query, frame-level matching and video-level matching may assign different ground-truth objects.
Beyond matching, $g$ and $h$ also differ from $f$ in what they add on top of its queries.
The tracker adds attention layers that, at each frame, fuse the current segmenter queries with the previous frame's tracked output. 
The refiner lets each tracked object exchange information in three directions: along its own trajectory (temporal attention and convolution), with the other objects (self-attention), and with the frozen per-frame segmenter queries (cross-attention). It then re-predicts class and mask from the whole trajectory.

The tracker adds learnable attention layers that, at each
frame, fuse the current segmenter queries with the previous frame's tracked
output. The refiner takes the tracker's identity-consistent object
embeddings and lets each object exchange information along its own
trajectory, through temporal self-attention and a short temporal convolution
within a track, self-attention across objects, and cross-attention to the
frozen per-frame embeddings, before re-predicting class and mask from the
whole trajectory. 

\paragraph{Hard-region supervision.}
The round-one segmenter serves as both the teacher that defines the hard
region and the initialisation for round two, which is then trained under a
fresh learning-rate schedule. The hard region is dilated by 5 pixels
before the loss is applied to it. The $\beta$-head duplicates
only the mask projection, an \mbox{MLP($256$, $256$, $256$)} of the same
shape as the original, and is dropped after training, so the model evaluated
in Section~\ref{subsec:results} has the same architecture and inference cost
as the baseline. At test time the two rounds are ensembled as $K = 2$.

\subsection{Results}
\label{subsec:results}
We begin with the challenge result on the test set, scored by the official
online evaluation system, and then decompose the
effectiveness of the proposed methods through ablations on the validation
set. VPS consists of semantic segmentation and tracking.
Section~\ref{ssub:semantic} takes the semantic half, separating the
contribution of the segmentation ensemble (SE, Section~\ref{sec:inference}) from that of
hard-region training on the segmenter (HRT, Section~\ref{subsec:hard-region}) and then asking which classes
the gain lands on. Section~\ref{ssub:vps} takes the full video panoptic
task, where identity across frames and across cameras also counts, and
isolates what decoupled inference merge (DIM, Section~\ref{sec:inference}) contributes;
it closes with the online setting, which we report to study the effect of
HRT on tracking, although the challenge submission did not use it. Unless
stated otherwise, the challenge result is on the test set and everything
that follows is on the $20$ validation videos.

\subsubsection{Challenge Result}

\begin{table}[htbp]
\centering
\begin{tabular}{lccc}
\toprule
method & wSTQ & wAQ & mIoU \\
\midrule
DVIS\_Plus\_M1\_NN\_fusion (ours) & \textbf{0.3547} & \textbf{0.2071} & \textbf{0.6075} \\
Pano-CTVPS                  & 0.3187 & 0.1781 & 0.5702 \\
Jay\_Street\_PVPS\_v0.1     & 0.2929 & 0.1486 & 0.5772 \\
AmapNet-v\_230523181741     & 0.2757 & 0.1466 & 0.5185 \\
AmapNet-v\_230501183044     & 0.2428 & 0.1039 & 0.5672 \\
ViP-DeepLab Baseline S      & 0.1750 & 0.1062 & 0.2883 \\
Clip KMax with Video Stitching & 0.0831 & 0.0251 & 0.2748 \\
\midrule
Our internal baseline$^{\dagger}$ & 0.3308 & 0.1929 & 0.5675 \\
\bottomrule
\end{tabular}
\caption{Waymo Open Dataset 2D Video Panoptic Segmentation leaderboard on the
test set, ordered by wSTQ. $\dagger$~our internal baseline, not a leaderboard
entry.}
\label{tab:leaderboard}
\end{table}

Table~\ref{tab:leaderboard} places our submission against the other entries on the
challenge test set. Our method leads on all three metrics, by $+0.0360$ wSTQ, $+0.0290$ wAQ
and $+0.0303$ mIoU over the next best entry on each metric. The last row is not a leaderboard entry: it
is the same pipeline built on the baseline segmenter rather than ours, evaluated on
the same test set, and it isolates what our contribution adds:$+0.0239$ wSTQ,
$+0.0142$ wAQ and $+0.0400$ mIoU.

\subsubsection{Semantic Segmentation}
\label{ssub:semantic}
Table~\ref{tab:leaderboard} reports the system as a whole.
Table~\ref{body:official-t1} separates its segmentation gain into the part
contributed by the segmentation ensemble (SE) and the part contributed by
hard-region training (HRT). We write HRT$_f$ for HRT applied to the
segmenter and HRT$_g$ for HRT applied to the tracker. This subsection, on
semantic segmentation, uses only HRT$_f$; the effect of HRT$_g$ is deferred
to the next subsection (Section~\ref{ssub:vps}), on panoptic segmentation.

\begin{table}[htbp]
\centering
\begin{tabular}{lccc}
\toprule
 & Baseline-segmenter & $+$ SE & $+$ SE, HRT$_f$ (ours) \\
\midrule
mIoU     & 0.6290 & 0.6353 & \textbf{0.6445} \\
$\Delta$ & ---    & $+0.0063$ & $+0.0092$ \\
\bottomrule
\end{tabular}
\caption{The baseline, the two-model ensemble, and the two-model ensemble
whose second model is trained with the hard-region loss. Each $\Delta$ is against
the column to its left, so the first isolates ensembling and the second isolates
the hard region.}
\label{body:official-t1}
\end{table}

Of the total $+0.0155$ mIoU, SE accounts for
$+0.0063$ and the HRT$_f$ accounts for the remaining $+0.0092$.
Neither subsumes the other: ensembling helps a model already trained with the
hard region, and the hard region helps a model that has already been ensembled.

Next, we study how that gain is distributed across the individual classes. Table~\ref{tab:per-class-rare} shows that our method
improves the rare classes $\mathcal{R}$, by about $2\%$ on average,
and the other frequent classes by about $1.5\%$ on average in mIoU. Note that
both methods are unable to detect ego vehicle and motorcyclist. Our hypothesis is
that motorcyclist is the least frequent object in the dataset, and that both
motorcyclist and ego vehicle have a high rate of mislabelling.

\begin{table}[htbp]
\centering
\begin{tabular}{lcccc}
\toprule
class & Baseline-segmenter & Ours & $\Delta$ & GT\% \\
\midrule
trailer            & 0.6432 & 0.6641 & $+0.0209$ & 0.022 \\
construction cone  & 0.5536 & 0.5875 & $+0.0339$ & 0.062 \\
ego vehicle        & 0.0000 & 0.0000 & $+0.0000$ & 0.004 \\
cyclist            & 0.8404 & 0.8583 & $+0.0179$ & 0.018 \\
motorcycle         & 0.7653 & 0.7828 & $+0.0175$ & 0.020 \\
bicycle            & 0.7319 & 0.7502 & $+0.0183$ & 0.028 \\
pedestrian object  & 0.2816 & 0.3009 & $+0.0193$ & 0.020 \\
motorcyclist       & 0.0000 & 0.0000 & $+0.0000$ & 0.000 \\
ground animal      & 0.1968 & 0.2071 & $+0.0103$ & 0.001 \\
bird               & 0.4912 & 0.5521 & $+0.0609$ & 0.000 \\
\midrule
$\mathcal{R}$ (10 classes)      & 0.4504 & 0.4703 & $+0.0199$ & 0.175 \\
complement (18 classes)         & 0.7193 & 0.7343 & $+0.0150$ & 99.826 \\
all (28 classes)                & 0.6233 & 0.6400 & $+0.0167$ & 100.000 \\
\bottomrule
\end{tabular}
\caption{Per-class IoU on the rare classes $\mathcal{R}$: the baseline
versus ours, the SE with the HRT$_f$-trained second
model. Classes are ordered by pixel count. \emph{GT\%} is the class's share
of all ground-truth pixels. The last three rows compare $\mathcal{R}$ with
its complement and with all 28 classes.}
\label{tab:per-class-rare}
\end{table}

\subsubsection{Video Panoptic Segmentation}
\label{ssub:vps}
So far we have compared the segmentation quality of the baseline and our
method, measured by mIoU, which scores each frame on its own. We now turn to
the final task, video panoptic segmentation, which adds identity: whether an
object keeps a single id across frames and across the five cameras.

DVIS++ adds two modules after the Mask2Former~\cite{mask2former} instance segmenter for VPS, the referring tracker and the temporal refiner, and trains them in a cascaded fashion, freezing the previous module when training the next one.
We run DVIS++'s VPS inference throughout, with the confidence threshold of Section~\ref{subsec:baseline} set to $\tau = 0.8$.
At inference, DVIS++ forms the final panoptic map from the two tracking modules, and the predictions of $f$, which provide the instance segmentation, are completely discarded.
Table~\ref{tab:official-val} shows what the full pipeline costs in semantic quality by adding ID association: the baseline scores $0.5919$ mIoU, against the $0.6290$ its own segmenter reaches under VSS inference.
We next shows the proposed DIM inference scheme for the VPS task, which re-includes the instance maps from $f$ in the inference pipeline and not only brings mIoU back but also improves wAQ.

\emph{Inputs and Effect of DIM.}
It takes two inputs: the semantic prediction from the segmenter $f$, and the panoptic prediction from the tracker $g$ and refiner $h$, produced exactly as in the baseline. In every result we report, including the competition submission, except the HRT$_g$ result in Table~\ref{tab:hr-tracking-online}, the panoptic prediction simply comes from the baseline, trained without any hard region training. HRT is used only for the segmenter $f$ that provides the semantic prediction.

Due to time and compute constraints, we did not carry HRT to the third stage of training, the refiner, and therefore simply use the baseline's panoptic prediction in DIM to produce the main results. To show the effectiveness of HRT for tracking as well, we conducted a separate experiment adding it to the second stage of training. Since this result is separate from the competition result, and we defer it to Table~\ref{tab:hr-tracking-online} as a special case.

\begin{table}[htbp]
\centering
\begin{tabular}{lccc}
\toprule
model & wSTQ & wAQ & mIoU \\
\midrule
baseline & 0.3379 & 0.1929 & 0.5919 \\
$+$DIM (ours)  & 0.3590 & 0.2049 & 0.6290 \\
$+$SE-HRT$_f$, DIM  (ours) & \textbf{0.3654} & \textbf{0.2071} & \textbf{0.6445} \\
\bottomrule
\end{tabular}
\caption{Official Waymo metric on the 20 validation videos, cross-camera
aligned, \textbf{offline setting}. \emph{baseline}: DVIS++ trained and run
as released. \emph{$+$DIM}: the baseline model with decoupled inference
merge at test time. \emph{$+$DIM (ours), HRT$_f$ (ours)}: the segmenter trained
with the hard-region loss and the restart ensemble, with DIM at test time.}
\label{tab:official-val}
\end{table}

In both DIM rows of Table~\ref{tab:official-val}, the panoptic map used for fusion is the baseline's, without any hard region training; SE-HRT$_f$ changes only the segmenter. Comparing the baseline with $+$DIM shows that using DIM at inference is essential for decoupled cascaded training: adding DIM alone improves wSTQ by 2.1 points. Adding SE-HRT$_f$ to the segmenter further improves wSTQ by 0.6 points, for a total gain of 2.8 points over the baseline.

\begin{figure}[t]
\centering
\includegraphics[width=\linewidth]{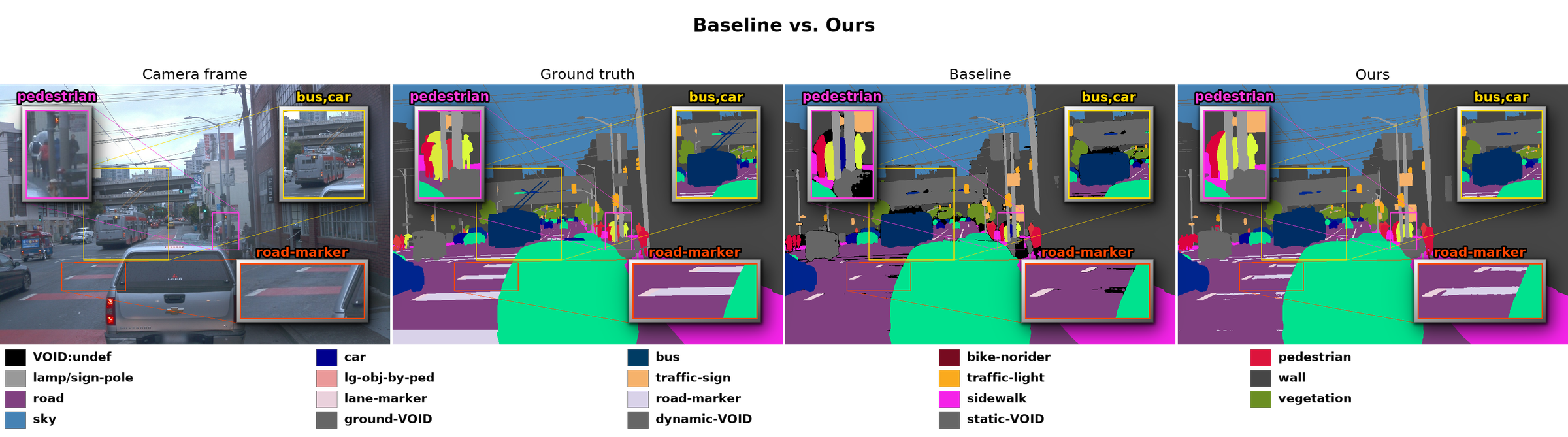}
\caption{The panoptic prediction on one frame of the front camera, coloured
by semantic class. \emph{Left to right:} the camera image, the ground truth,
the baseline, and ours; three regions---a pedestrian group, a bus and car,
and a road marker---are magnified beside each panel, with the class legend
below. Black is \texttt{VOID:undef}, which a method emits where no retained
query claims the pixel; it can appear only under the VPS readout of
Section~\ref{subsec:baseline}, never under VSS. The baseline leaves large
\texttt{VOID:undef} regions over the vehicles and across the pedestrian
group, where ours commits to a class.}
\label{fig:method-comparison}
\end{figure}

Figure~\ref{fig:method-comparison} shows a qualitative comparison on one frame. Compared with the baseline, our method better detects several critical objects in this frame: cars, buses, pedestrians, and road signs.

\paragraph{Special Case: Online}
Video panoptic segmentation is usually evaluated in two settings that differ
in the temporal horizon available at inference. In the online setting, frames
are processed in order, and the prediction for frame $t$ may use only frames
up to $t$. In the offline setting, the whole video is available before
prediction, so frame $t$ may use frames both before and after it. For the
competition, we submitted results in the offline setting, since the
competition provides offline access and the longer horizon gives better
results than the online setting. In DVIS++, the online model is the
segmenter with the referring tracker, which realizes the online constraint
recurrently by taking the current frame's queries and the previous frame's
tracked queries as its only inputs.
In this setting, the pipeline stops at the referring tracker and uses its output as the panoptic prediction. 
Here, we additionally evaluate the effectiveness of adding hard region supervision to the training of the referring tracker (HRS$_g$). The results in Table~\ref{tab:hr-tracking-online} demonstrate that adding HRT$_g$ to the training of later modules also helps VPS: it improves wSTQ by 1.3 points on its own and by 1.1 points on top of SE-HRT$_f$ and DIM, and together with our other methods, it improves wSTQ by 5.1 points over the online baseline.

\begin{table}[htbp]
\centering
\begin{tabular}{lccc}
\toprule
method & wSTQ & wAQ & mIoU \\
\midrule
baseline                    & 0.2712 & 0.1297 & 0.5669 \\
$+$HRT$_g$ (ours)                  & 0.2843 & 0.1449 & 0.5579 \\
\midrule
$+$SE-HRT$_f$,DIM (ours)               & 0.3113 & 0.1503 & 0.6445 \\
$+$HRT$_g$, SE-HRT$_f$, DIM (ours) & \textbf{0.3225} & \textbf{0.1614} & \textbf{0.6445} \\
\bottomrule
\end{tabular}
\caption{Hard-region loss on the referring tracker (HRL$_T$) and decoupled
inference, \textbf{online setting}; the temporal refiner is not used.
Each ``$+$'' row adds the named component to the baseline. Best wSTQ in bold.}
\label{tab:hr-tracking-online}
\end{table}

Table~\ref{tab:hr-tracking-online} compares four methods in the online setting. \emph{Baseline} uses the baseline alone. \emph{+HRT$_g$} adds the hard-region loss to the tracker's training. \emph{+SE-HRT$_f$, DIM} adds the decoupled inference merge of the segmenter and tracker outputs, along with the hard region training of the segmenter. \emph{+HRT$_g$, SE-HRT$_f$, DIM} (\emph{Ours}) combines all three.
Comparing \emph{Baseline} with \emph{+HRT$_g$} shows that adding the hard-region loss alone improves wAQ by 1.5 points, and the last two rows show that this gain persists after DIM and SE-HRT$_f$ (+1.1 points). Note that the same comparison shows that adding the hard-region loss drops mIoU (by 0.9 points). Our hypothesis is that, when training the tracker, the major gain is in ID association between consecutive frames, and AQ is a direct metric of association quality, hence the AQ gain. This result is interesting: when the model is trained for the VPS objective, yet takes as input queries learned for instance segmentation, a trade-off forms between association and segmentation quality. When AQ is the main goal, the method improves AQ. It also decreases mIoU. From another perspective, this is consistent with the flaw of decoupled, cascaded training discussed in Section~\ref{par:cascade-flaw}. DIM addresses this problem at inference time, along with hard region training at training time: \emph{Ours} recovers mIoU to 0.6445 and improves wSTQ by 5.1 points over \emph{Baseline}.

\ifdefined\NOAPPENDIX\else
\paragraph{Cross-camera alignment}
Figure~\ref{fig:overlap-objects} in Appendix~\ref{app:cross-cam} shows the overlap objects recovered by Algorithm~\ref{alg:cross-cam} on a randomly chosen frame, next to the ground truth; the two are very close.
\fi

\section{Limitations and Future Work}

For the segmenter in our VPS pipeline, our improvements come from hard-region training and ensembling the resulting models at inference time. While this boosting paradigm improves segmentation quality, it multiplies inference cost by the number of ensemble members, which is impractical for self-driving, where real-time inference is required. A natural next step is to collapse the boosting pipeline into a single model via knowledge distillation, training one student to match the ensemble's outputs~\cite{hinton2015distilling}. Prior work shows that a single student can recover most of an ensemble's accuracy~\cite{furlanello2018born}, and that it can additionally retain the ensemble's uncertainty estimates~\cite{malinin2020ensemble}t. Allen-Zhu and Li~\cite{allenzhu2020towards} further provide a theoretical account of why a single model can absorb the ensemble's gains. Whether this holds for hard-region-specialized models, whose members are deliberately trained on different data distributions, remains to be verified.

The second problem we address in the baseline is information loss from cascade training in the decoupled modules: the segmenter, tracker, and refiner are trained sequentially toward different stage-wise objectives, each with its predecessor frozen, and each later module reads only queries, never pixel features, when forming its queries, so pixels lost at one stage cannot be recovered by the next.  DIM recovers these pixels by re-including the segmenter's instance maps at inference via nearest-neighbor merging, but this ensembles two modules and relies on KNN relabeling, both impractical for self-driving. The next step is a single end-to-end model, in one of two directions. The first is to let later stages re-attend to the pixel features rather than only to the previous stage's queries. The second is to close the objective gap between stages, a known problem in staged optimization addressed by continuation methods~\cite{mobahi2015link}, progressively stricter cascades~\cite{cai2018cascade}, and residual schemes that train each new stage on the gap left by the frozen predecessor~\cite{friedman2001greedy,fahlman1990cascade}. 

\clearpage
\bibliographystyle{plainnat}
\bibliography{refs}

\ifdefined\NOAPPENDIX\else
\clearpage
\section{Appendix}

Three things are deferred here. Section~\ref{app:cross-cam} gives the
qualitative check on cross-camera alignment, Section~\ref{app:per-class} the
full per-class breakdown behind the rare-class table, and
Section~\ref{app:miou-drop} the decomposition of the refiner's mIoU drop into
the part contributed by training and the part contributed by the readout.

\subsection{Cross-camera alignment}
\label{app:cross-cam}
Figure~\ref{fig:overlap-objects} qualitatively shows the overlap objects recovered by Algorithm~\ref{alg:cross-cam} on a randomly chosen frame, next to the ground-truth overlap objects. The two lower rows are very close: each vehicle straddling a camera boundary is recovered as one object with a single global id, except for a few very small objects (some of which are visibly the same object across two cameras but are not associated in the ground truth, which gives them different ids) and one object that the prediction mis-associates.

\begin{figure}[htbp]
\centering
\includegraphics[width=\linewidth]{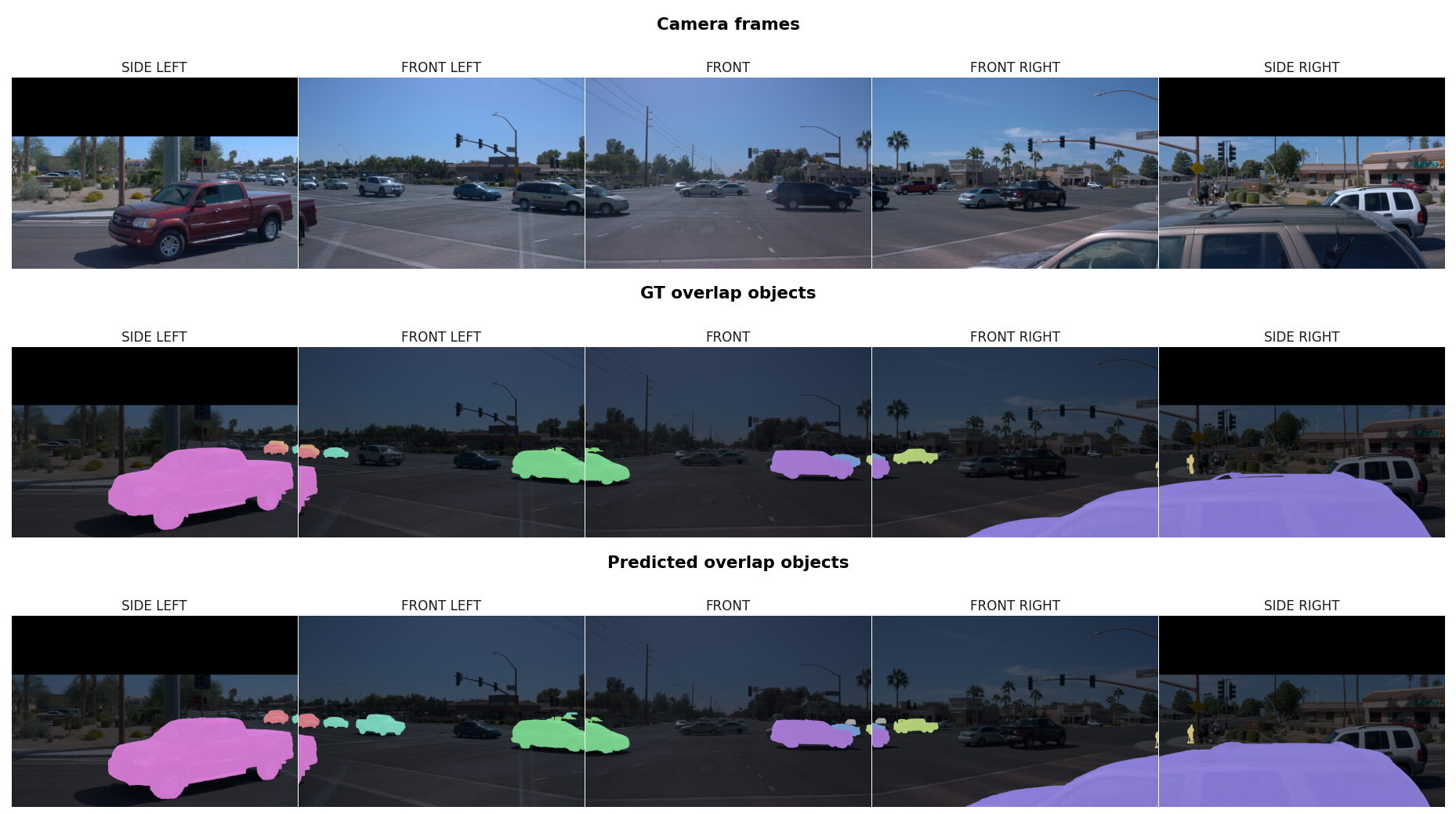}
\caption{Cross-camera overlap objects on one frame. \emph{Top:} the five camera images, arranged left to right in panoramic order. \emph{Middle:} the ground-truth overlap objects. \emph{Bottom:} the overlap objects recovered by Algorithm~\ref{alg:cross-cam}, that is, the objects found to span more than one camera; each is shaded by its global id, so an object keeps a single color across every camera it appears in. The two lower rows use matched colors, so an object present in both appears in the same color in each.}
\label{fig:overlap-objects}
\end{figure}

\subsection{Per-class Segmentation Results}
\label{app:per-class}

\begin{table}[htbp]
\centering
\footnotesize
\setlength{\tabcolsep}{3pt}
\resizebox{\linewidth}{!}{%
\begin{tabular}{lr ccc ccc ccc}
\toprule
 & & \multicolumn{3}{c}{IoU} & \multicolumn{3}{c}{Precision} & \multicolumn{3}{c}{Recall} \\
\cmidrule(lr){3-5} \cmidrule(lr){6-8} \cmidrule(lr){9-11}
Class & GT \% & Baseline & Ours & $\Delta$ & Baseline & Ours & $\Delta$ & Baseline & Ours & $\Delta$ \\
\midrule
truck & 0.53 & 0.5970 & 0.7110 & +0.1140 & 0.6964 & 0.8147 & +0.1183 & 0.8071 & 0.8481 & +0.0410 \\
bird & 0.00 & 0.4877 & 0.5468 & +0.0592 & 0.6326 & 0.7317 & +0.0991 & 0.6804 & 0.6840 & +0.0036 \\
cone & 0.04 & 0.5524 & 0.5860 & +0.0337 & 0.6257 & 0.6479 & +0.0222 & 0.8251 & 0.8598 & +0.0347 \\
lane-marker & 0.44 & 0.6493 & 0.6755 & +0.0262 & 0.8016 & 0.8254 & +0.0238 & 0.7736 & 0.7881 & +0.0145 \\
road-marker & 0.60 & 0.6233 & 0.6492 & +0.0259 & 0.8271 & 0.8345 & +0.0074 & 0.7167 & 0.7451 & +0.0284 \\
bike-norider & 0.03 & 0.7369 & 0.7588 & +0.0219 & 0.8707 & 0.8854 & +0.0147 & 0.8275 & 0.8414 & +0.0139 \\
dynamic-VOID & 0.31 & 0.6447 & 0.6663 & +0.0216 & 0.8096 & 0.8429 & +0.0333 & 0.7600 & 0.7607 & +0.0007 \\
traffic-light & 0.14 & 0.7235 & 0.7449 & +0.0214 & 0.8733 & 0.8921 & +0.0188 & 0.8083 & 0.8186 & +0.0103 \\
motorcycle & 0.02 & 0.7693 & 0.7906 & +0.0213 & 0.8748 & 0.8971 & +0.0223 & 0.8644 & 0.8694 & +0.0050 \\
cyclist & 0.02 & 0.8447 & 0.8631 & +0.0184 & 0.9292 & 0.9422 & +0.0130 & 0.9028 & 0.9113 & +0.0085 \\
lg-obj-by-ped & 0.02 & 0.2873 & 0.3042 & +0.0169 & 0.5073 & 0.5813 & +0.0740 & 0.3985 & 0.3895 & $-$0.0090 \\
lamp/sign-pole & 0.87 & 0.6848 & 0.7016 & +0.0168 & 0.8451 & 0.8564 & +0.0113 & 0.7830 & 0.7952 & +0.0122 \\
ground-VOID & 4.61 & 0.7746 & 0.7896 & +0.0150 & 0.8694 & 0.8855 & +0.0161 & 0.8766 & 0.8794 & +0.0028 \\
trailer & 0.02 & 0.6587 & 0.6730 & +0.0142 & 0.8613 & 0.8402 & $-$0.0211 & 0.7369 & 0.7718 & +0.0349 \\
sidewalk & 6.04 & 0.8512 & 0.8615 & +0.0103 & 0.9226 & 0.9272 & +0.0046 & 0.9166 & 0.9240 & +0.0074 \\
car & 9.34 & 0.9377 & 0.9480 & +0.0103 & 0.9710 & 0.9728 & +0.0018 & 0.9647 & 0.9737 & +0.0090 \\
ground-animal & 0.00 & 0.2057 & 0.2156 & +0.0099 & 0.6349 & 0.6907 & +0.0558 & 0.2332 & 0.2386 & +0.0054 \\
pedestrian & 0.59 & 0.8663 & 0.8757 & +0.0094 & 0.9197 & 0.9260 & +0.0063 & 0.9372 & 0.9416 & +0.0044 \\
static-VOID & 3.70 & 0.6262 & 0.6344 & +0.0082 & 0.7764 & 0.7914 & +0.0150 & 0.7639 & 0.7618 & $-$0.0021 \\
traffic-sign & 0.38 & 0.6933 & 0.6990 & +0.0057 & 0.8106 & 0.8209 & +0.0103 & 0.8274 & 0.8248 & $-$0.0026 \\
road & 19.96 & 0.9396 & 0.9435 & +0.0039 & 0.9672 & 0.9681 & +0.0009 & 0.9705 & 0.9738 & +0.0033 \\
wall & 25.22 & 0.9146 & 0.9175 & +0.0029 & 0.9523 & 0.9524 & +0.0001 & 0.9584 & 0.9615 & +0.0031 \\
vegetation & 14.24 & 0.8903 & 0.8926 & +0.0023 & 0.9437 & 0.9455 & +0.0018 & 0.9402 & 0.9410 & +0.0008 \\
sky & 12.88 & 0.9446 & 0.9449 & +0.0003 & 0.9708 & 0.9701 & $-$0.0007 & 0.9722 & 0.9732 & +0.0010 \\
ego veh & 0.00 & 0.0000 & 0.0000 & +0.0000 & 0.0000 & 0.0000 & +0.0000 & 0.0000 & 0.0000 & +0.0000 \\
motorcyclist & 0.00 & 0.0000 & 0.0000 & +0.0000 & 0.0000 & 0.0000 & +0.0000 & 0.0000 & 0.0000 & +0.0000 \\
\midrule
bus & 0.01 & 0.4480 & 0.4378 & $-$0.0103 & 0.5513 & 0.5583 & +0.0070 & 0.7053 & 0.6696 & $-$0.0357 \\
other-lg-veh & 0.01 & 0.2595 & 0.2158 & $-$0.0437 & 0.3369 & 0.3418 & +0.0049 & 0.5304 & 0.3692 & $-$0.1612 \\
\bottomrule
\end{tabular}}
\caption{Per-class IoU, precision and recall on the validation split: the baseline against ours, the two-model ensemble with the hard-region second model. Classes are ordered by change in IoU; the two below the rule are the only ones whose IoU the ensemble degrades. \emph{GT \%} is each class's share of the ground-truth pixels.}
\label{app:official-t2}
\end{table}

Table~\ref{app:official-t2} gives the full per-class breakdown behind Table~\ref{tab:per-class-rare}. In the main body, we report the rare classes $\mathcal{R}$ individually, since they are the hardest to detect, and average the remaining classes. Here all 28 classes are listed.

IoU improves on 24 of the 28 classes, is unchanged on two (ego vehicle and motorcyclist, which neither model detects), and drops on two. For 15 of the 24 improved classes, precision gains more than recall, so the second model mainly removes false positives. The largest gain is on truck (+11.4 IoU), driven by precision (+11.8). The frequent, large-area classes (road, wall, vegetation, sky) gain less than 0.4 IoU, as they are already near their ceiling.

The two classes that drop, bus and other-large-vehicle, both lose recall ($-$3.6 and $-$16.1 points) while their precision slightly rises. Both are visually close to truck, which gains the most. Our hypothesis is that some of their pixels are now assigned to truck.

\subsection{Decomposing the Refiner's mIoU Drop: Training vs.\ Readout}
\label{app:miou-drop}

In the main body, when we take the output from the refiner, we follow DVIS++ and use VPS inference to produce the panoptic map, and we report the mIoU of this panoptic map. Here, we separate the refiner's mIoU loss into two comparisons. First, we fix the readout to VSS for both modules (Table~\ref{tab:vss-inference}), so that only semantic segmentation quality is measured. Even without the VPS readout, cascaded training alone lowers mIoU by 1.6 points, from 0.6290 for the segmenter to 0.6130 for the refiner. The two are trained with the same loss but matched differently: the segmenter per frame, the refiner per video. Video-level matching assigns each query one ground-truth track and one class for the whole clip. For stuff, a track is simply the class region, which is stable over time, so this assignment fits every frame; for a thing object, a single assignment can fit some frames poorly. The cascade cannot correct this by design: the refiner forms its queries only from upstream queries, which were trained under frame-level matching, and never attends to pixel features, so a frame the video-level assignment fits poorly cannot be recovered from the image.

\begin{table}[htbp]
\centering
\begin{tabular}{lc}
\toprule
model & mIoU \\
\midrule
baseline-segmenter & 0.6290 \\
baseline-refiner   & 0.6130 \\
\bottomrule
\end{tabular}
\caption{Video semantic segmentation (VSS) inference for both modules, on the 20 validation videos.}
\label{tab:vss-inference}
\end{table}

\begin{table}[htbp]
\centering
\begin{tabular}{lcccc}
\toprule
Group & \#classes & VPS & VSS & $\Delta$ \\
\midrule
Thing & 8  & 0.5294 & 0.5515 & $+$0.0222 \\
Stuff & 20 & 0.6465 & 0.6376 & $-$0.0089 \\
\midrule
All   & 28 & 0.5919 & 0.6130 & $+$0.0211 \\
\bottomrule
\end{tabular}
\caption{mIoU of the same refiner under VPS and VSS inference, on the 20 validation videos. $\Delta$ is VSS minus VPS.}
\label{tab:vps-vss}
\end{table}

With the model fixed to the refiner, the readout shifts quality between thing and stuff classes (Table~\ref{tab:vps-vss}): VSS is 2.2 points better on thing classes, and VPS is 0.9 points better on stuff classes. VPS keeps a query only if its confidence exceeds $\tau$ and assigns each pixel to a single winning query, so thing objects predicted with low confidence lose their pixels. VSS applies no threshold and sums every query's contribution at each pixel, so low-confidence thing queries still contribute, while stuff regions receive probability mass from thing queries along their borders.

\fi

\end{document}